%% file: paper.tex
\documentclass{article}

\usepackage[nonatbib,preprint]{neurips_2020}
\usepackage[numbers]{natbib}

\usepackage[utf8]{inputenc} % allow utf-8 input
\usepackage[T1]{fontenc}    % use 8-bit T1 fonts
\usepackage{hyperref}       % hyperlinks
\usepackage{url}            % simple URL typesetting
\usepackage{booktabs}       % professional-quality tables
\usepackage{amsfonts}       % blackboard math symbols
\usepackage{nicefrac}       % compact symbols for 1/2, etc.
\usepackage{microtype}      % microtypography
\usepackage{xcolor}
\usepackage{xspace}
\usepackage{tabularx}
\usepackage{amsmath,bm,multicol,multirow}
\usepackage{cleveref}
\usepackage[pdftex]{graphicx}
\usepackage{array, makecell}
\usepackage[scientific-notation=true]{siunitx}
\usepackage{subcaption}
\usepackage{chngpage}

\usepackage{ulem}
\input{macro}

\title{\LARGE \wvpp{}: A Framework for Self-Supervised Learning of Speech Representations}

\author{%
  Alexei Baevski \And Henry Zhou \And Abdelrahman Mohamed \And Michael Auli \AND 
  {\tt \{abaevski,henryzhou7,abdo,michaelauli\}@fb.com} \\\AND
  Facebook AI
}

\begin{document}

\maketitle

\begin{abstract}
  We show for the first time that learning powerful representations from speech audio alone followed by fine-tuning on transcribed speech can outperform the best semi-supervised methods while being conceptually simpler.
  wav2vec 2.0 masks the speech input in the latent space and solves a contrastive task defined over a quantization of the latent representations which are jointly learned.
  Experiments using all labeled data of Librispeech achieve 1.8/3.3 WER on the clean/other test sets.
  When lowering the amount of labeled data to one hour, wav2vec 2.0 outperforms the previous state of the art on the 100 hour subset while using 100 times less labeled data.
  Using just ten minutes of labeled data and pre-training on 53k hours of unlabeled data still achieves 4.8/8.2 WER.
  This demonstrates the feasibility of speech recognition with limited amounts of labeled data.\footnote{Code and models are available at \url{https://github.com/pytorch/fairseq}}
\end{abstract}

\section{Introduction}

Neural networks benefit from large quantities of labeled training data. 
However, in many settings labeled data is much harder to come by than unlabeled data: 
current speech recognition systems require thousands of hours of transcribed speech to reach acceptable performance which is not available for the vast majority of the nearly 7,000 languages spoken worldwide~\citep{lewis2016ethnologue}. 
Learning purely from labeled examples does not resemble language acquisition in humans: infants learn language by listening to adults around them - a process that requires learning good representations of speech.

In machine learning, self-supervised learning has emerged as a paradigm to learn general data representations from unlabeled examples and to fine-tune the model on labeled data. 
This has been particularly successful for natural language processing~\citep{peters2018acl,radford2018unsup,devlin2018bert} and is an active research area for computer vision~\citep{henaff2019cpc,bachman2019learning,misra2019selfsupervised,he2019momentum,chen2020simple}.

In this paper, we present a framework for self-supervised learning of representations from raw audio data. 
Our approach encodes speech audio via a multi-layer convolutional neural network and then masks spans of the resulting latent speech representations~\citep{jiang2019improving,wang2020unsupervised}, similar to masked language modeling~\citep{devlin2018bert}.
The latent representations are fed to a Transformer network to build contextualized representations and the model is trained via a contrastive task where the true latent is to be distinguished from distractors~\citep{oord2018cpc,schneider2019wav2vec,rivire2020unsupervised,kawakami2020learning} ~(\autoref{sec:model}).

\begin{figure}[t]
\centering
\includegraphics[width=0.7\linewidth]{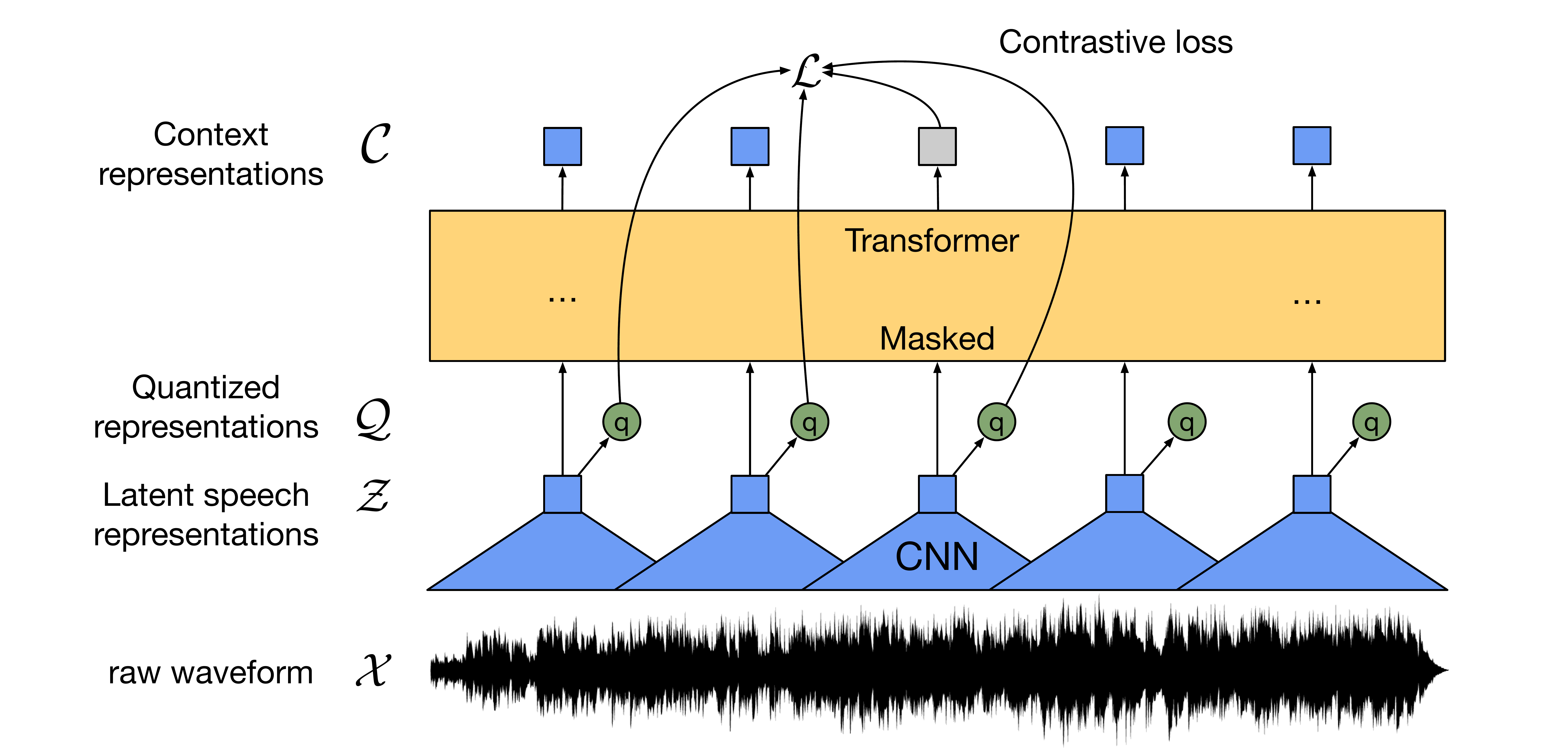}
\caption{Illustration of our framework which jointly learns contextualized speech representations and an inventory of discretized speech units.}
\label{fig:illustration}
\end{figure}

As part of training, we learn discrete speech units~\citep{oord2017neural,alex2019unsupervised,chorowski2019unsup,harwath2019learning} via a gumbel softmax~\citep{jang2016gumbel,baevski2019vqwav2vec} to represent the latent representations in the contrastive task (\autoref{fig:illustration}) which we find to be more effective than non-quantized targets.
After pre-training on unlabeled speech, the model is fine-tuned on labeled data with a Connectionist Temporal Classification (CTC) loss~\citep{graves2006ctc,baevski2019effectiveness} to be used for downstream speech recognition tasks (\autoref{sec:training})

Previous work learned a quantization of the data followed by a contextualized representations with a self-attention model~\citep{baevski2019vqwav2vec,baevski2019effectiveness}, whereas our approach solves both problems end-to-end.
Masking parts of the input with Transformer networks for speech has been explored ~\citep{baevski2019effectiveness,jiang2019improving}, but prior work relies either on a two-step pipeline or their model is trained by reconstructing the filter bank input features.
Other related work includes learning representations from auto-encoding the input data~\citep{tjandra2019vqvae,eloff2019unsupervised} or directly predicting future timesteps~\citep{chung2019apc}.

Our results show that jointly learning discrete speech units with contextualized representations achieves substantially better results than fixed units learned in a prior step~\cite{baevski2019effectiveness}.
We also demonstrate the feasibility of ultra-low resource speech recognition: 
when using only 10 minutes of labeled data, our approach achieves word error rate (WER) 4.8/8.2 on the clean/other test sets of \libri{}.
We set a new state of the art on TIMIT phoneme recognition as well as the 100 hour clean subset of \libri{}. 
Moreover, when we lower the amount of labeled data to just one hour, we still outperform the previous state of the art self-training method of~\citep{park2020improved} while using 100 times less labeled data and the same amount of unlabeled data.
When we use all 960 hours of labeled data from \libri{}, then our model achieves 1.8/3.3 WER (\autoref{sec:setup}, \autoref{sec:results}).

\section{Model}
\label{sec:model}

Our model is composed of a multi-layer convolutional feature encoder $f: \Inp \mapsto \Feat$ which takes as input raw audio $\Inp$ and outputs latent speech representations $\ze_1, \dots, \ze_T$ for $T$ time-steps. 
They are then fed to a Transformer $g: \Feat \mapsto \Context$ to build representations $\cc_1, \dots, \cc_T$ capturing  information from the entire sequence~\cite{devlin2018bert,baevski2019vqwav2vec,baevski2019effectiveness}.
The output of the feature encoder is discretized to $\zq_t$ with a quantization module $\Feat \mapsto \QFeat$ to represent the targets (\autoref{fig:illustration}) in the self-supervised objective (\autoref{sec:objective}).
Compared to vq-wav2vec~\citep{baevski2019vqwav2vec}, our model builds context representations over continuous speech representations and self-attention captures dependencies over the entire sequence of latent representations end-to-end.

\paragraph{Feature encoder.} 
The encoder consists of several blocks containing a temporal convolution followed by layer normalization~\citep{ba2016layer} and a GELU activation function~\citep{hendrycks2016gaussian}.
The raw waveform input to the encoder is normalized to zero mean and unit variance.
The total stride of the encoder determines the number of time-steps $T$ which are input to the Transformer (\autoref{sec:setup-pretrain}).

\paragraph{Contextualized representations with Transformers.}
The output of the feature encoder is fed to a context network which follows the Transformer architecture~\cite{vaswani2017transformer,devlin2018bert,liu2019roberta}. 
Instead of fixed positional embeddings which encode absolute positional information, we use a convolutional layer similar to~\cite{mohamed2019libri,baevski2019effectiveness,wu2019pay} which acts as relative positional embedding.
We add the output of the convolution followed by a GELU to the inputs and then apply layer normalization.

\paragraph{Quantization module.} 
For self-supervised training we discretize the output of the feature encoder $\ze$ to a finite set of speech representations via product quantization~\citep{jegou2011ieee}.
This choice led to good results in prior work which learned discrete units in a first step followed by learning contextualized representations~\cite{baevski2019vqwav2vec}.
Product quantization amounts to choosing quantized representations from multiple codebooks and concatenating them. 
Given $G$ codebooks, or groups, with $V$ entries $e \in \R^{V \times d/G}$, we choose one entry from each codebook and concatenate the resulting vectors $e_1, \dots, e_G$ and apply a linear transformation $\R^d \mapsto \R^f$ to obtain $\zq \in \R^f$.

The Gumbel softmax enables choosing discrete codebook entries in a fully differentiable way~\citep{gumbel1954statistical,jang2016gumbel,maddison2014sampling}. 
We use the straight-through estimator~\citep{jiang2019improving} and setup $G$ hard Gumbel softmax operations~\citep{jang2016gumbel}. 
The feature encoder output $\ze$ is mapped to $\lll \in \R^{G \times V}$ logits and the probabilities for choosing the $v$-th codebook entry for group $g$ are
\begin{equation}
    p_{g,v} = \frac{\exp(l_{g,v} + n_v)/\tau}{\sum_{k = 1}^{V} \exp(l_{g,k} + n_k)/\tau}, %\quad v_j \iid \Gumbel(0,1),
\label{eq:gumbel}
\end{equation}
where $\tau$ is a non-negative temperature, $n = -\log(-\log(u))$ and $u$ are uniform samples from $\mathcal{U}(0, 1)$.
During the forward pass, codeword $i$ is chosen by $i = \text{argmax}_j p_{g,j}$ and in the backward pass, the true gradient of the Gumbel softmax outputs is used.

\section{Training}
\label{sec:training}

To pre-train the model we mask a certain proportion of time steps in the latent feature encoder space (\autoref{sec:masking}), similar to masked language modeling in BERT~\citep{devlin2018bert}.
The training objective requires identifying the correct quantized latent audio representation in a set of distractors for each masked time step (\autoref{sec:objective}) and the final model is fine-tuned on the labeled data (\autoref{sec:train_finetune}).

\subsection{Masking}
\label{sec:masking}
We mask a proportion of the feature encoder outputs, or time steps before feeding them to the context network and replace them with a trained feature vector shared between all masked time steps;
we do not mask inputs to the quantization module.
To mask the latent speech representations output by the encoder, we randomly sample without replacement a certain proportion $p$ of all time steps to be starting indices and then mask the subsequent $M$ consecutive time steps from every sampled index; spans may overlap.

\subsection{Objective}
\label{sec:objective}

During pre-training, we learn representations of speech audio by solving a contrastive task $\Lmlm{}$ which requires to identify the true quantized latent speech representation for a masked time step within a set of distractors.
This is augmented by a codebook diversity loss $\Ld{}$ to encourage the model to use the codebook entries equally often. 
\begin{equation}
    \mathcal{L} = \Lmlm{} + \alpha \Ld{}
\label{eq:loss}
\end{equation}
where $\alpha$ is a tuned hyperparameter.

\paragraph{Contrastive Loss.}
Given context network output $\cc_t$ centered over masked time step $t$, the model needs to identify the true quantized latent speech representation $\zq_t$ in a set of $K+1$ quantized candidate representations $\zqt \in \mathbf{Q}_t$ which includes $\zq_t$ and $K$ distractors~\citep{gutmann2010aistats,oord2018cpc}.
Distractors are uniformly sampled from other masked time steps of the same utterance. The loss is defined as
\begin{equation}
    \Lmlm{} = -\log \frac{\exp(sim(\cc_t, \zq_t)/\kappa)}{\sum_{\zqt \sim \mathbf{Q}_t} \exp(sim(\cc_t, \zqt)/\kappa)}
\label{eq:mlm}
\end{equation}
where we compute the cosine similarity $sim(\mathbf{a},\mathbf{b}) = \mathbf{a}^{T} \mathbf{b} / \|\mathbf{a}\| \|\mathbf{b}\|$ between context representations and quantized latent speech representations~\citep{he2019momentum,chen2020simple}.

\paragraph{Diversity Loss.}
The contrastive task depends on the codebook to represent both positive and negative examples and the diversity loss $\mathcal{L}_d$ is designed to increase the use of the quantized codebook representations~\citep{dieleman2018challenge}.
We encourage the equal use of the $V$ entries in each of the $G$ codebooks by maximizing the entropy of the averaged softmax distribution $\lll$ over the codebook entries for each codebook $\bar{p}_{g}$ across a batch of utterances; the softmax disribution does not contain the gumbel noise nor a temperature:\footnote{Our implementation maximizes perplexity $\frac{GV - \sum_{g=1}^{G}\exp(-\sum_{v=1}^{V}p_{gv} \log{p_{gv}})}{GV}$ which is equivalent.}
\begin{equation}
    \Ld{} = \frac{1}{GV} \sum_{g=1}^{G} - H(\bar{p}_{g}) = \frac{1}{GV} \sum_{g=1}^{G} \sum_{v=1}^{V} \bar{p}_{g,v} \log{\bar{p}_{g,v}}
\end{equation}

\subsection{Fine-tuning}
\label{sec:train_finetune}

Pre-trained models are fine-tuned for speech recognition by adding a randomly initialized linear projection on top of the context network into $C$ classes representing the vocabulary of the task~\citep{baevski2019effectiveness}.
For \libri{}, we have 29 tokens for character targets plus a word boundary token. 
Models are optimized by minimizing a CTC loss~\citep{graves2006ctc} and we apply a modified version of SpecAugment~\citep{park2019specaugment} by masking to time-steps and channels during training which delays overfitting and significantly improves the final error rates, especially on the \libril{} subsets with few labeled examples.

\section{Experimental Setup}
\label{sec:setup}

\subsection{Datasets}
\label{sec:setup-datasets}

As unlabeled data we consider the \libri{} corpus~\citep{panayotov2015librispeech} without transcriptions containing 960 hours of audio (\librisz{}) or the audio data from \vox{} (\voxsz{}). 
For the latter we follow the pre-processing of~\citep{kahn2020librilight} resulting in 53.2k hours of audio.
We fine-tune on five labeled data settings: 
960 hours of transcribed \libri{}, the train-clean-100 subset comprising 100 hours (100 hours labeled), as well as the \libril{} limited resource training subsets originally extracted from \libri{}, these are train-10h (10 hours labeled), train-1h (1 hour labeled), train-10min (10 min labeled). 
We follow the evaluation protocol of \libril{} for these splits and evaluate on the standard Librispech dev-other/clean and test-clean/other sets.

We fine-tune the pre-trained models for phoneme recognition on the TIMIT dataset~\citep{garofolo1993timit}. 
It contains five hours of audio recordings with detailed phoneme labels.
We use the standard train, dev and test split and follow the standard protocol of collapsing phone labels to 39 classes.

\subsection{Pre-training}
\label{sec:setup-pretrain}

Models are implemented in fairseq~\cite{ott2019fairseq}.
For masking, we sample $p=0.065$ of all time-steps to be starting indices and mask the subsequent $M=10$ time-steps.
This results in approximately 49\% of all time steps to be masked with a mean span length of 14.7, or 299ms (see~\autoref{app:masking} for more details on masking).

The feature encoder contains seven blocks and the temporal convolutions in each block have 512 channels with strides (5,2,2,2,2,2,2) and kernel widths (10,3,3,3,3,2,2).
This results in an encoder output frequency of 49 hz with a stride of about 20ms between each sample, and a receptive field of 400 input samples or 25ms of audio. 
The convolutional layer modeling relative positional embeddings has kernel size 128 and 16 groups.

We experiment with two model configurations which use the same encoder architecture but differ in the Transformer setup: \wvppbase{} contains 12 transformer blocks, model dimension 768, inner dimension (FFN) 3,072 and 8 attention heads. 
Batches are built by cropping 250k audio samples, or 15.6sec, from each example. 
Crops are batched together to not exceed 1.4m samples per GPU and we train on a total of 64 V100 GPUs for 1.6 days~\cite{ott2018scaling}; the total batch size is 1.6h.

The \wvppbig{} model contains 24 transformer blocks with model dimension 1,024, inner dimension 4,096 and 16 attention heads. 
We crop 320k audio samples, or 20sec, with a limit of 1.2m samples per GPU and train on 128 V100 GPUs over 2.3 days for \libri{} and 5.2 days for \vox{}; the total batch size is 2.7h.
We use dropout 0.1 in the Transformer, at the output of the feature encoder and the input to the quantization module. 
Layers are dropped at a rate of 0.05 for \wvppbase{} and 0.2 for \wvppbig{}~ \cite{huang2016deep,fan2019reducing}; there is no layer drop for \voxsz{}.

We optimize with Adam~\citep{kingma2015adam}, warming up the learning rate for the first 8\% of updates to a peak of $\Enot{5e-4}$ for \wvppbase{} and $\Enot{3e-4}$ for \wvppbig{}, and then linearly decay it. 
\wvppbig{} trains for 250k updates, \wvppbase{} for 400k updates, and \wvppbig{} on \voxsz{} for 600k updates.
We use weight $\alpha=0.1$ for the diversity loss~\autoref{eq:loss}.
For the quantization module we use $G=2$ and $V=320$ for both models, resulting in a theoretical maximum of 102.4k codewords. 
Entries are of size $d/G = 128$ for \wvppbase{} amd $d/G = 384$ for \wvppbig{}.
The Gumbel softmax temperature $\tau$ is annealed from 2 to a minimum of 0.5 for~\wvppbase{} and 0.1 for~\wvppbig{} by a factor of 0.999995 at every update. 
The temperature in the contrastive loss (\autoref{eq:mlm}) is set to $\kappa=0.1$.
For the smaller Librispeech dataset, we regularize the model by applying an L2 penalty to the activations of the final layer of the feature encoder and scale down the gradients for the encoder by a factor of 10. 
We also use a slightly different encoder architecture where we do not use layer normalization, and instead of normalizing the raw waveform, the output of the first encoder layer is normalized.
In the contrastive loss we use $K=100$ distractors.
We choose the training checkpoint with the lowest $\Lmlm{}$ on the validation set.

\subsection{Fine-tuning}
\label{sec:setup-finetune}

After pre-training we fine-tune the learned representations on labeled data and add a randomly initialized output layer on top of the Transformer to predict characters (\libri{}/\libril{}) or phonemes (TIMIT).
For \libril{}, we train three seeds with two different learning rates (2e-5 and 3e-5) for all subsets and choose the configuration with lowest WER on dev-other subset decoded with the official 4-gram language model (LM) with beam 50 and fixed model weights (LM weight 2, word insertion penalty -1). 
For \wvppbase{} on the labeled 960h subset we use a learning rate of 1e-4.

We optimize with Adam and a tri-state rate schedule where the learning rate is warmed up for the first 10\% of updates, held constant for the next 40\% and then linearly decayed for the remainder. 
\wvppbase{} uses a batch size of 3.2m samples per GPU and we fine-tune on 8 GPUs, giving a total batch size of 1,600sec. 
\wvppbig{} batches 1.28m samples on each GPU and we fine-tune on 24 GPUs, resulting in an effective batch size of 1,920sec.
For the first 10k updates only the output classifier is trained, after which the Transformer is also updated. 
The feature encoder is not trained during fine-tuning. 
We mask the feature encoder representations with a strategy similar to SpecAugment~\citep{park2019specaugment} detailed in~\autoref{app:finetune}.

\subsection{Language Models and Decoding}

We consider two types of language models (LM): a 4-gram model and a Transformer~\citep{baevski2018adaptive} trained on the \libri{} LM corpus.
The Transformer LM is identical to~\citep{synnaeve2020end} and contains 20 blocks, model dimension 1,280, inner dimension 6,144 and 16 attention heads. 
We tune the weights of the language model (interval $[0,5]$) and a word insertion penalty ($[-5,5]$) via Bayesian optimization\footnote{\url{https://github.com/facebook/Ax}}: 
we run 128 trials with beam 500 for the 4-gram LM and beam 50 for the Transformer LM and choose the best set of weights according to performance on dev-other.
Test performance is measured with beam 1,500 for the n-gram LM and beam 500 for the Transformer LM.
We use the beam search decoder of~\cite{pratap2019w2l}.

\section{Results}
\label{sec:results}

\begin{table}[t]
\caption{WER on the \libri{} dev/test sets when training on the \libril{} low-resource labeled data setups of 10 min, 1 hour, 10 hours and the clean 100h subset of \libri{}. 
Models use either the audio of \libri{} (\librisz{}) or the larger \vox{} (\voxsz{}) as unlabeled data.
We consider two model sizes: \wvppbase{} (95m parameters) and \wvppbig{} (317m parameters).
Prior work used 860 unlabeled hours (LS-860) but the total with labeled data is 960 hours and comparable to our setup.
}
\label{tab:librilight}
\centering 
\begin{tabular}{lccrrrrr}
\toprule
\multirow{2}{*}{Model} & Unlabeled & \multirow{2}{*}{LM} & \multicolumn{2}{c}{dev} && \multicolumn{2}{c}{test} \\
\cline{4-5}\cline{7-8} 
{} & data & {} & clean & other && clean & other \\
\midrule
\midrule
\multicolumn{8}{l}{\textbf{10 min labeled}}\\
Discrete BERT~\citep{baevski2019effectiveness} & \librisz{} & 4-gram & 15.7 & 24.1 && 16.3 & 25.2 \\
% \cmidrule{1-1}
\midrule
% \textbf{This work} \\
\wvppbase{} & \librisz{} & 4-gram & 8.9 & 15.7 && 9.1 & 15.6 \\
&& Transf. & 6.6 & 13.2 && 6.9 & 12.9 \\
\wvppbig{} & \librisz{} & Transf. & 6.6 & 10.6 && 6.8 & 10.8 \\
& \voxsz{} & Transf. & 4.6 & 7.9 && 4.8 & 8.2 \\
\midrule
\midrule
\multicolumn{8}{l}{\textbf{1h labeled}}\\
Discrete BERT~\citep{baevski2019effectiveness} & \librisz{} & 4-gram & 8.5 & 16.4 && 9.0 & 17.6 \\
% \cmidrule{1-1}
\midrule
% \textbf{This work} \\
\wvppbase{} & \librisz{} & 4-gram & 5.0 & 10.8 && 5.5 & 11.3 \\
&& Transf. & 3.8 & 9.0 && 4.0 & 9.3 \\
\wvppbig{} & \librisz{} & Transf. & 3.8 & 7.1 && 3.9 & 7.6 \\
& \voxsz{} & Transf. & 2.9 & 5.4 && 2.9 & 5.8 \\
\midrule
\midrule
\multicolumn{8}{l}{\textbf{10h labeled}}\\
Discrete BERT~\citep{baevski2019effectiveness} & \librisz{} & 4-gram & 5.3 & 13.2 && 5.9 & 14.1 \\
Iter. pseudo-labeling~\cite{xu2020iterative} & \librisz{} & 4-gram+Transf. & 23.51 & 25.48 && 24.37 & 26.02 \\
{} & \voxsz{} & 4-gram+Transf. & 17.00 & 19.34 && 18.03 & 19.92 \\
% \cmidrule{1-1}
\midrule
% \textbf{This work} \\
\wvppbase{} & \librisz{} & 4-gram & 3.8 & 9.1 && 4.3 & 9.5 \\
&& Transf. & 2.9 & 7.4 && 3.2 & 7.8 \\
\wvppbig{} & \librisz{} & Transf. & 2.9 & 5.7 && 3.2 & 6.1 \\
& \voxsz{} & Transf. & 2.4 & 4.8 && 2.6 & 4.9 \\
\midrule
\midrule
\multicolumn{8}{l}{\textbf{100h labeled}}\\
Hybrid DNN/HMM \citep{L_scher_2019} & - & 4-gram & 5.0 & 19.5 && 5.8 & 18.6 \\
TTS data augm.~\citep{aleks2020need} & - & LSTM & & && 4.3 & 13.5 \\
Discrete BERT~\citep{baevski2019effectiveness} & \librisz{} & 4-gram & 4.0 & 10.9 && 4.5 & 12.1 \\
Iter. pseudo-labeling~\cite{xu2020iterative} & \libriunsz{} & 4-gram+Transf. & 4.98 & 7.97 && 5.59 & 8.95 \\
& \voxsz{} & 4-gram+Transf. & 3.19 & 6.14 && 3.72 & 7.11 \\
Noisy student~\citep{park2020improved} & \libriunsz{} & LSTM & 3.9 & 8.8 && 4.2 & 8.6 \\
% \cmidrule{1-1}
\midrule
% \textbf{This work} \\
\wvppbase{} & \librisz{} & 4-gram & 2.7 & 7.9 && 3.4 & 8.0 \\
&& Transf. & 2.2 & 6.3 && 2.6 & 6.3 \\
\wvppbig{} & \librisz{} & Transf. & 2.1 & 4.8 && 2.3 & 5.0 \\
& \voxsz{} & Transf. & 1.9 & 4.0 && 2.0 & 4.0 \\
\bottomrule
\end{tabular}
\end{table}

\subsection{Low-Resource Labeled Data Evaluation}

We first evaluate our pre-trained models in settings where the amount of labeled data is limited to get a sense of how the representations learned on unlabeled data can improve low resource settings.
If a pre-trained model captures the structure of speech, then it should require few labeled examples to fine-tune it for speech recognition.
The models are pre-trained on the audio data of either \libri{} (\librisz{}) or \vox{} (\voxsz{}) and most results are obtained by decoding with a Transformer language model (Transf.); \autoref{app:libri} shows results with no language model at all as well as with an n-gram language model.

The \wvppbig{} model pre-trained on \voxsz{} and fine-tuned on only 10 minutes of labeled data achieves a word error rate of 5.2/8.6 on the \libri{} clean/other test sets.
Ten minutes of labeled data corresponds to just 48 recordings with an average length of 12.5 seconds.
This demonstrates that ultra-low resource speech recognition is possible with self-supervised learning on unlabeled data.
Our approach of jointly learning discrete units and contextualized representations clearly improves over previous work which learned quantized audio units in a separate step~\citep{baevski2019effectiveness}, reducing WER by a about a third.

A recent iterative self-training approach~\citep{park2020improved} represents the state of the art on the clean 100 hour subset of \libri{} but it requires multiple iterations of labeling, filtering, and re-training. 
Our approach is simpler: we pre-train on the unlabeled data and fine-tune on the labeled data.
On the 100 hour subset of Librispeech, their method achieves WER 4.2/8.6 on test-clean/other which compares to WER 2.3/5.0 with the \wvppbig{} model in a like for like setup, a relative WER reduction of 45\%/42\%.

When the \wvppbig{} model uses an order of magnitude less labeled data (10h labeled), then it still achieves WER 3.2/6.1, an error reduction of 24\%/29\% relative to iterative self-training.
Using only a single hour of labeled data, the same model achieves WER 3.9/7.6 which improves on both test-clean and test-other by 7\%/12\% - with two orders of magnitude less labeled data.
We note that the \libril{} data splits contain both clean and noisy data leading to better accuracy on test-other compared to test-clean.
Increasing model size reduces WER on all setups with the largest improvements on test-other (\wvppbase{} vs. \wvppbig{} both on \librisz{}) and increasing the amount of unlabeled training data also leads to large improvements (\wvppbig{} \librisz{} vs. \voxsz{}).

\subsection{High-Resource Labeled Data Evaluation on \libri{}}

In this section we evaluate the performance when large quantities of labeled speech are available to assess the effectiveness of our approach in a high resource setup.
Specifically, we fine-tune the same models as before on the full 960 hours of labeled \libri{}: \wvppbase{} and \wvppbig{} pre-trained on \librisz{} as well as \wvppbig{} pre-trained on \voxsz{}.

\autoref{tab:librispeech} shows that our approach achieves WER 1.8/3.3 on test-clean/other on the full Librispeech benchmark.
This is despite a weaker baseline architecture: 
supervised training of our architecture achieves WER 2.1/4.6 (\wvppbig{} - from scratch) compared to WER 1.9/4.1 for ContextNet~\citep{han2020contextnet}, the baseline architecture of the state of the art~\citep{park2020improved}.
We use a simple Transformer with CTC which does not perform as well as seq2seq models~\citep{synnaeve2020end}. 

Note that the vocabulary of our acoustic model (characters) does not match the vocabulary of the LM (words) which delays feedback from the LM and is likely to be detrimental. 
Most recent work~\citep{synnaeve2020end,xu2020iterative,han2020contextnet,park2020improved} uses the better performing word pieces~\citep{schuster2012wordpieces} for both models.
Moreover, our result is achieved without any data balancing such as \cite{park2020improved}.
Finally, self-training is likely complimentary to pre-training and their combination may yield even better results.
\autoref{app:analysis_errors} presents a detailed error analysis of our pre-trained models in various labeled data setups.

\begin{table}
\caption{%
WER on \libri{} when using all 960 hours of labeled data (cf. \autoref{tab:librilight}).
}
\label{tab:librispeech}
\centering 
\begin{tabular}{lccrrrrr}
\toprule
\multirow{2}{*}{Model} & Unlabeled & \multirow{2}{*}{LM} & \multicolumn{2}{c}{dev} && \multicolumn{2}{c}{test} \\
\cline{4-5}\cline{7-8} 
{} & data & {} & clean & other && clean & other \\
\midrule
\textbf{Supervised} \\
CTC Transf~\citep{synnaeve2020end} & - & CLM+Transf. & 2.20 & 4.94 && 2.47 & 5.45 \\
S2S Transf. ~\citep{synnaeve2020end} & - & CLM+Transf. & 2.10 & 4.79 && 2.33 & 5.17 \\
Transf. Transducer~\citep{zhang2020transformer} & - & Transf. & - & - && 2.0 & 4.6 \\
ContextNet~\citep{han2020contextnet} & - & LSTM & 1.9 & 3.9 && 1.9 & 4.1 \\
Conformer~\citep{gulati2020conformer} & - & LSTM & 2.1 & 4.3 && 1.9 & 3.9 \\
\midrule
\textbf{Semi-supervised} \\
CTC Transf. + PL~\citep{synnaeve2020end} & \voxsz{} & CLM+Transf. &  2.10 & 4.79 && 2.33 & 4.54 \\
S2S Transf. + PL~\citep{synnaeve2020end} & \voxsz{} & CLM+Transf. &  2.00 & 3.65 && 2.09 & 4.11 \\
Iter. pseudo-labeling~\cite{xu2020iterative} & \voxsz{} & 4-gram+Transf. & 1.85 & 3.26 && 2.10 & 4.01 \\
Noisy student~\citep{park2020improved} & \voxsz{} & LSTM & 1.6 & 3.4 && 1.7 & 3.4 \\
\midrule
\textbf{This work} \\
\wvppbig{} - from scratch & - & Transf. & 1.7 & 4.3 && 2.1 & 4.6 \\
\wvppbase{} & \librisz{} & Transf. & 1.8 & 4.7 && 2.1 & 4.8 \\
\wvppbig{} & \librisz{} & Transf. & 1.7 & 3.9 && 2.0 & 4.1 \\
& \voxsz{} & Transf. & 1.6 & 3.0 && 1.8 & 3.3 \\
\bottomrule
\end{tabular}
\end{table}

\subsection{Phoneme Recognition on TIMIT}

Next, we evaluate accuracy on TIMIT phoneme recognition by fine-tuning the pre-trained models on the labeled TIMIT training data. 
We fine-tune as for the 10 hour subset of \libril{} but do not use a language model.
\autoref{tbl:timit-results} shows that our approach can achieve a new state of the art on this dataset, reducing PER by a relative 23\%/29\% over the next best result on the dev/test sets.
\autoref{app:analysis_latents} shows an analysis of how the discrete latent speech representations related to phonemes.
Other recent work on pre-training which evaluates on TIMIT includes~\citep{ravanelli2020pasep} who solve multiple tasks to learn good representations of speech.

\begin{table}
\caption{%
TIMIT phoneme recognition accuracy in terms of phoneme error rate (PER).
}
\label{tbl:timit-results}
\centering
\begin{tabular}{lrr}
\toprule
{} &   dev PER &  test PER \\
\midrule
CNN + TD-filterbanks~\citep{zeghidour2018filters} & 15.6 & 18.0 \\
PASE+~\citep{ravanelli2020pasep} & - & 17.2 \\
Li-GRU + fMLLR~\citep{ravanelli2018light} & -- & 14.9 \\ %$\pm$ 0.27 \\
wav2vec~\citep{schneider2019wav2vec} & {12.9} & {14.7}  \\
vq-wav2vec~\citep{baevski2019vqwav2vec} & 9.6 & 11.6 \\
\midrule
\textbf{This work (no LM)} \\ 
\wvppbig{} (\librisz{}) & 7.4 & 8.3 \\
\bottomrule
\end{tabular}
\end{table}

\subsection{Ablations}
\label{sec:ablations}

A difference to previous work~\cite{baevski2019vqwav2vec,baevski2019effectiveness} is that we quantize the latent audio representations only for the contrastive loss, i.e., when latents are used as \emph{targets}, but not when the latents are \emph{input} to the Transformer network. 
We motivate this choice by an ablating for which we adopt a reduced training setup to increase experimental turn around: 
we pre-train \wvppbase{} on \librisz{} for 250k updates with masking probability $p=0.075$, fine-tune on train-10h for 60k updates on a single GPU with 640k samples per batch, or 40 sec of speech audio.
We report the average WER and standard deviation on the concatenation of dev-clean and dev-other (dev PER) for three seeds of fine-tuning.

Table \ref{tbl:quantization} shows that our strategy of continuous inputs with quantized targets (Baseline) performs best. 
Continuous latent speech representations retain more information to enable better context representations and quantizing the target representations leads to more robust training.
Quantizing the latents both in the input and the targets performs least well, and explains the lower performance of prior work~\citep{baevski2019vqwav2vec,baevski2019effectiveness}.
Continuous targets reduce the effectiveness of self-supervised training since targets can capture detailed artifacts of the current sequence, e.g. speaker and background information, which make the task easier and prevent the model from learning general representations beneficial to speech recognition.
The training accuracy of identifying the correct latent audio representation increases from 62\% to 78.0\% when switching from quantized to continuous targets.
Continuous inputs and continuous targets perform second best but various attempts to improve it did not lead to better results (see \autoref{app:ablations} for this experiment and other ablations on various hyperparameters).

\begin{table}[t]
\caption{%
Average WER and standard deviation on combined dev-clean/other of \libri{} for three training seeds. We ablate quantizing the context network input and the targets in the contrastive loss.
}
\label{tbl:quantization}
\centering
\begin{tabular}{lrrr}
\toprule
{} & avg. WER & std. \\
\midrule
Continuous inputs, quantized targets (Baseline) & 7.97 & 0.02  \\
% \midrule
Quantized inputs, quantized targets & 12.18 & 0.41 \\
Quantized inputs, continuous targets & 11.18 & 0.16 \\
Continuous inputs, continuous targets & 8.58 & 0.08 \\
\bottomrule
\end{tabular}
\end{table}

\section{Conclusion}

We presented \wvpp{}, a framework for self-supervised learning of speech representations which masks latent representations of the raw waveform and solves a contrastive task over quantized speech representations.
Our experiments show the large potential of pre-training on unlabeled data for speech processing: 
when using only 10 minutes of labeled training data, or 48 recordings of 12.5 seconds on average, we achieve a WER of 4.8/8.2 on test-clean/other of \libri{}. 

Our model achieves results which achieve a new state of the art on the full Librispeech benchmark for noisy speech.
On the clean 100 hour \libri{} setup, wav2vec 2.0 outperforms the previous best result while using 100 times less labeled data.
The approach is also effective when large amounts of labeled data are available. 
We expect performance gains by switching to a seq2seq architecture and a word piece vocabulary.

\section*{Broader Impact}

There are around 7,000 languages in the world and many more dialects. 
However, for most of them no speech recognition technology exists since current systems require hundreds or thousands of hours of labeled data which is hard to collect for most languages.
We have shown that speech recognition models can be built with very small amounts of annotated data at very good accuracy.
We hope our work will make speech recognition technology more broadly available to many more languages and dialects.

\begin{ack}
We thank Tatiana Likhomanenko and Qiantong Xu for helpful discussion and their help with wav2letter integration.

\end{ack}

\bibliography{main}

\bibliographystyle{abbrvnat}

\newpage

\appendix
\renewcommand{\thesection}{\Alph{section}}

\section*{Appendices}

\section{Masking distribution}
\label{app:masking}

When choosing which time-steps to mask, each latent speech representation in an utterance is considered a candidate starting time-step with probability $p$ where $M$ is the length of each masked span starting from the respective time step; both are hyper-parameters.
Sampled starting time steps are expanded to length $M$ and spans can overlap.

\begin{figure}[h]
\centering
\includegraphics[width=0.4\linewidth]{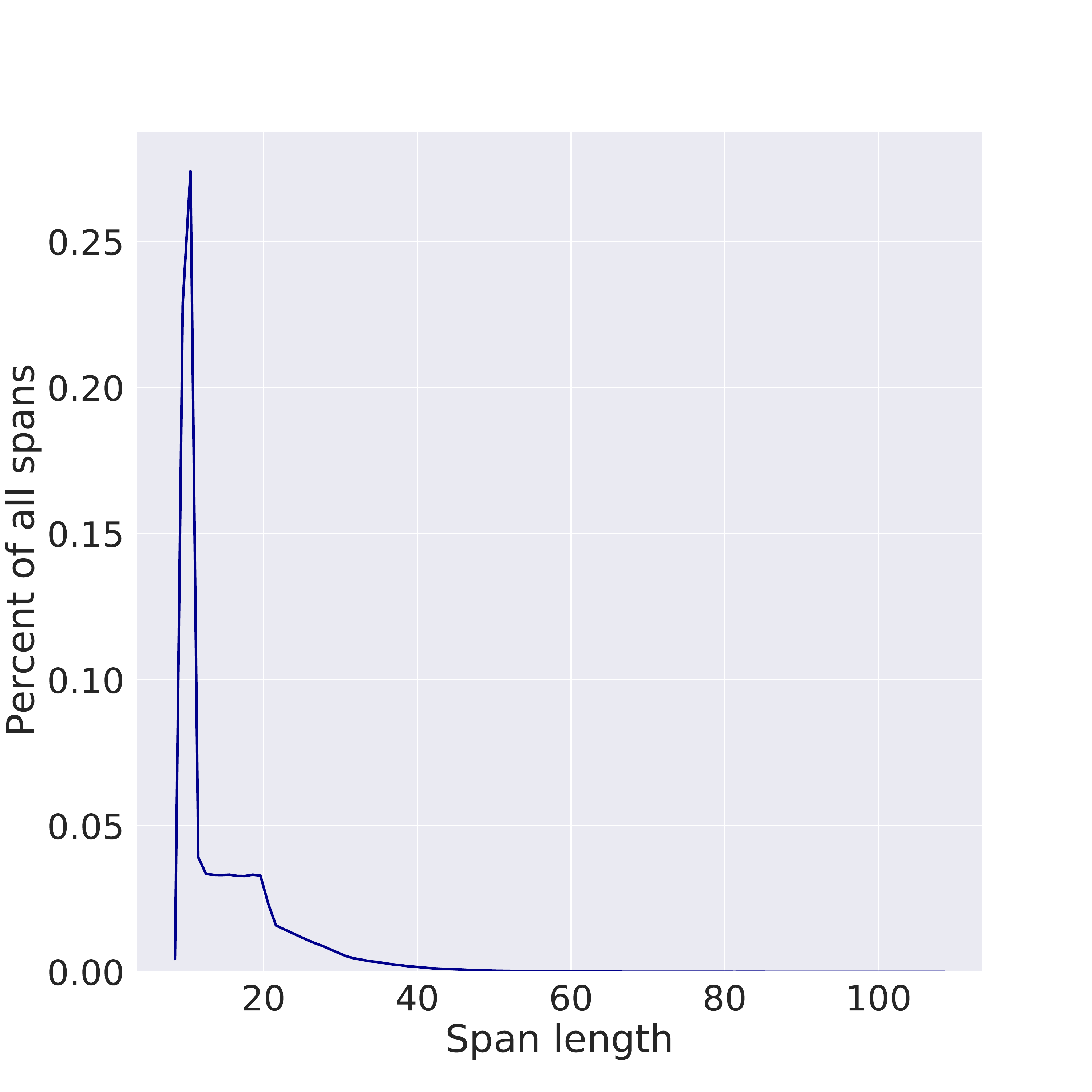}
\caption{Mask length distribution for a 15 second sample with $p=0.065$ and $M=10$.
}
\label{fig:masks}
\end{figure}

For a 15 sec long audio sample, the average mask length is 14.7 time-steps, corresponding to 299ms of audio, with a median of 10 time-steps, and a maximum of about 100 time steps; about 49\% of all time-steps in the sample will be masked. 
A plot of the corresponding mask length distribution is shown in~\autoref{fig:masks} and an ablation of $M$ and $p$ as well as the effect of other masking strategies is shown in~\autoref{tbl:masks}. 
Reducing $M$ results in increased prediction accuracy for the self-supervised but the task becomes trivial when spans with length one are masked, leading to poor performance on downstream speech recognition tasks.
We also consider other masking strategies: w/o overlap uniform($a$,$b$) samples for each starting index a span length $M^s$ from interval $a$ to $b$ and masks the subsequent $M^s$ time-steps taking care not to overlap with existing spans;
poisson($\lambda$) and normal($\mu$, $\sigma$) sample $M^s$ from Poisson and normal distributions.

\begin{table}[h]
\caption{%
Ablations on settings for the masking strategy during pre-training. 
When masking without overlap, we choose starting time steps with $p=0.037$ which results in the total number of masked tokens to match the baseline.
}
\label{tbl:masks}
\centering
\begin{tabular}{lrr}
\toprule
{} & avg WER &  std \\
\midrule
Baseline ($p=0.075$) & 7.97 & 0.02 \\
\midrule
Mask length $M=8$ & 8.33 & 0.05 \\
Mask length $M=12$ & 8.19 & 0.08 \\
Mask length $M=15$ & 8.43 & 0.19 \\
\midrule
Mask probability $p=0.065$ & 7.95 & 0.08 \\
Mask probability $p=0.06$ & 8.14 & 0.22 \\
\midrule
Mask w/o overlap, uniform(1,31) & 8.39 & 0.02 \\
Mask w/o overlap, uniform(10,30) & 9.17 & 0.05 \\
Mask w/o overlap, poisson(15) & 8.13 & 0.04 \\
Mask w/o overlap, normal(15, 10) & 8.37 & 0.03 \\
Mask w/o overlap, length 10 & 9.15 & 0.02 \\
Mask w/o overlap, length 15 & 9.43 & 0.26 \\
\bottomrule
\end{tabular}
\end{table}

\section{Fine-tuning Setup}
\label{app:finetune}

During fine-tuning we apply a masking strategy to the feature encoder outputs similar to SpecAugment~\citep{park2019specaugment}:
we randomly choose a number of starting time steps for which a span of ten subsequent time-steps is replaced with a mask embedding; spans may overlap and we use the same masked time step embedding as during pre-training. 
We also mask channels by choosing a number of channels as starting indices and then expand each one to cover the subsequent 64 channels.
Spans may overlap and the selected channel spans are set to zero value.
We use LayerDrop~\citep{huang2016deep,fan2019reducing} at a rate of 0.05 for \wvppbase{} and 0.1 for \wvppbig{} during fine-tuning.

Table \ref{tbl:fine-tuning} summarizes the fine-tuning hyper-parameter settings used for the different labeled data setup.
Table \ref{tbl:decoding-params-libri} shows the decoding parameters used for final evaluations of the various labeled data setups for Librispeech pre-trained models and Table~\ref{tbl:decoding-params-vox} shows decoding parameters for \vox{}.

\begin{table}[ht!]
\caption{%
Fine-tuning hyperparameters
}
\label{tbl:fine-tuning}
\centering 
\begin{tabular}{lrrr}
\toprule
{} & timestep mask prob. & channel mask prob. & updates \\
\midrule
10 min & 0.075 & 0.008 & 12k \\
1 hour & 0.075 & 0.004 & 13k \\
10 hours & 0.065 & 0.004 & 20k \\
100 hours & 0.05 & 0.008 & 50k \\
960 hours & 0.05 & 0.0016 & 320k \\
TIMIT & 0.065 & 0.012 & 40k \\
\bottomrule
\end{tabular}
\end{table}

\begin{table}[ht!]
\caption{%
Decoding parameters for \libri{} subsets for models pre-trained on Librispeech
}
\label{tbl:decoding-params-libri}
\centering 
\begin{tabular}{lrrrr}
\toprule
{} & 4gram LM weight & 4gram word insert. & TransLM weight & TransLM word insert. \\
\midrule
10 min & 3.23 & -0.26 & 1.20 & -1.39 \\
1 hour & 2.90 & -1.62 & 1.15 & -2.08 \\
10 hours & 2.46 & -0.59  & 1.06 & -2.32 \\
100 hours & 2.15 & -0.52 & 0.87 & -1.00 \\
960 hours & 1.74 & 0.52 & 0.92 & -0.86 \\
\bottomrule
\end{tabular}
\end{table}

\begin{table}[ht!]
\caption{%
Decoding parameters for \libri{} subsets for models pre-trained on Librivox.
}
\label{tbl:decoding-params-vox}
\centering 
\begin{tabular}{lrrrr}
\toprule
{} & 4gram LM weight & 4gram word insert. & TransLM weight & TransLM word insert. \\
\midrule
10 min & 3.86 & -1.18 & 1.47 & -2.82 \\
1 hour & 3.09 & -2.33 & 1.33 & -0.69 \\
10 hours & 2.12 & -0.90  & 0.94 & -1.05 \\
100 hours & 2.15 & -0.52 & 0.87 & -1.00 \\
960 hours & 1.57 & -0.64 & 0.90 & -0.31 \\
\bottomrule
\end{tabular}
\end{table}

\newpage
\section{Full results for \libril{} and \libri{}}
\label{app:libri}

\begin{table}[h]
\caption{WER on the \libri{} dev/test sets when training on the \libril{} low-resource labeled data setups (cf.~\autoref{tab:librilight}).
}
\label{tab:librilight_full}
\centering 
\begin{tabular}{lcccrrrr}
\toprule
\multirow{2}{*}{Model} & Unlabeled & \multirow{2}{*}{LM} & \multicolumn{2}{c}{dev} && \multicolumn{2}{c}{test} \\
\cline{4-5}\cline{7-8} 
{} & data & {} & clean & other && clean & other \\
\midrule
\midrule
\multicolumn{8}{l}{\textbf{10 min labeled}}\\
\wvppbase{} & \librisz{} & None & 46.1 & 51.5 && 46.9 & 50.9 \\
&& 4-gram & 8.9 & 15.7 && 9.1 & 15.6 \\
&& Transf. & 6.6 & 13.2 && 6.9 & 12.9 \\
\wvppbig{} & \librisz{} & None & 43.0 & 46.3 && 43.5 & 45.3 \\
&& 4-gram & 8.6 & 12.9 && 8.9 & 13.1 \\
&& Transf. & 6.6 & 10.6 && 6.8 & 10.8 \\
\wvppbig{} & \voxsz{} & None & 38.3 & 41.0 && 40.2 & 38.7 \\
&& 4-gram & 6.3 & 9.8 && 6.6 & 10.3 \\
&& Transf. & 4.6 & 7.9 && 4.8 & 8.2 \\
\midrule
\midrule
\multicolumn{8}{l}{\textbf{1h labeled}}\\
\wvppbase{} & \librisz{} & None & 24.1 & 29.6 && 24.5 & 29.7 \\
&& 4-gram & 5.0 & 10.8 && 5.5 & 11.3 \\
&& Transf. & 3.8 & 9.0 && 4.0 & 9.3 \\
\wvppbig{} & \librisz{} & None & 21.6 & 25.3 && 22.1 & 25.3 \\
&& 4-gram & 4.8 & 8.5 && 5.1 & 9.4 \\
&& Transf. & 3.8 & 7.1 && 3.9 & 7.6 \\
\wvppbig{} & \voxsz{} & None & 17.3 & 20.6 && 17.2 & 20.3 \\
&& 4-gram & 3.6 & 6.5 && 3.8 & 7.1 \\
&& Transf. & 2.9 & 5.4 && 2.9 & 5.8 \\
\midrule
\midrule
\multicolumn{8}{l}{\textbf{10h labeled}}\\
\wvppbase{} & \librisz{} & None & 10.9 & 17.4 && 11.1 & 17.6 \\
&& 4-gram & 3.8 & 9.1 && 4.3 & 9.5 \\
&& Transf. & 2.9 & 7.4 && 3.2 & 7.8 \\
\wvppbig{} & \librisz{} & None & 8.1 & 12.0 && 8.0 & 12.1 \\
&& 4-gram & 3.4 & 6.9 && 3.8 & 7.3 \\
&& Transf. & 2.9 & 5.7 && 3.2 & 6.1 \\
\wvppbig{} & \voxsz{} & None & 6.3 & 9.8 && 6.3 & 10.0 \\
&& 4-gram & 2.6 & 5.5 && 3.0 & 5.8 \\
&& Transf. & 2.4 & 4.8 && 2.6 & 4.9 \\
\midrule
\midrule
\multicolumn{8}{l}{\textbf{100h labeled}}\\
\wvppbase{} & \librisz{} & None & 6.1 & 13.5 && 6.1 & 13.3 \\
&& 4-gram & 2.7 & 7.9 && 3.4 & 8.0 \\
&& Transf. & 2.2 & 6.3 && 2.6 & 6.3 \\
\wvppbig{} & \librisz{} & None & 4.6 & 9.3 && 4.7 & 9.0 \\
&& 4-gram & 2.3 & 5.7 && 2.8 & 6.0 \\
&& Transf. & 2.1 & 4.8 && 2.3 & 5.0 \\
\wvppbig{} & \voxsz{} & None & 3.3 & 6.5 && 3.1 & 6.3 \\
&& 4-gram & 1.8 & 4.5 && 2.3 & 4.6 \\
&& Transf. & 1.9 & 4.0 && 2.0 & 4.0 \\
\bottomrule
\end{tabular}
\end{table}

\begin{table}[h!]
\caption{%
WER on \libri{} when using all 960 hours of \libri{} as labeled data (cf. \autoref{tab:librispeech}).
}
\label{tab:librispeech_full}
\centering 
\begin{tabular}{lcccrrrr}
\toprule
\multirow{2}{*}{Model} & Unlabeled & \multirow{2}{*}{LM} & \multicolumn{2}{c}{dev} && \multicolumn{2}{c}{test} \\
\cline{4-5}\cline{7-8} 
{} & data & {} & clean & other && clean & other \\
\midrule
\wvppbig{} - from scratch & - & None & 2.8 & 7.6 && 3.0 & 7.7 \\
& - & 4-gram & 1.8 & 5.4 && 2.6 & 5.8 \\
& - & Transf. & 1.7 & 4.3 && 2.1 & 4.6 \\
\midrule
\wvppbase{} & \librisz{} & None & 3.2 & 8.9 && 3.4 & 8.5 \\
&& 4-gram & 2.0 & 5.9 && 2.6 & 6.1 \\
&& Transf. & 1.8 & 4.7 && 2.1 & 4.8 \\
\wvppbig{} & \librisz{} & None & 2.6 & 6.5 && 2.8 & 6.3 \\
&& 4-gram & 1.7 & 4.6 && 2.3 & 5.0 \\
&& Transf. & 1.7 & 3.9 && 2.0 & 4.1 \\
\wvppbig{} & \voxsz{} & None & 2.1 & 4.5 && 2.2 & 4.5 \\
&& 4-gram & 1.4 & 3.5 && 2.0 & 3.6 \\
&& Transf. & 1.6 & 3.0 && 1.8 & 3.3 \\
\bottomrule
\end{tabular}
\end{table}

\newpage 
\section{Analysis of Discrete Latent Speech Representations}
\label{app:analysis_latents}

Next, we investigate whether the discrete latent speech representations $\zq_t$ learned by the quantizer relate to phonetic information:
Using \wvppbig{} pre-trained on \voxsz{} and without any fine-tuning, we compute the discrete latents for the training data of TIMIT and compute the co-occurrence between human annotated phonemes and the latents.
Ties are broken by choosing the phoneme which is most represented in the receptive field of $\zq_t$.
The training data contains 3696 utterances of average length 13.6 sec, or 563k discrete latents.

\begin{figure}[h!]
\centering
\includegraphics[width=1.0\linewidth]{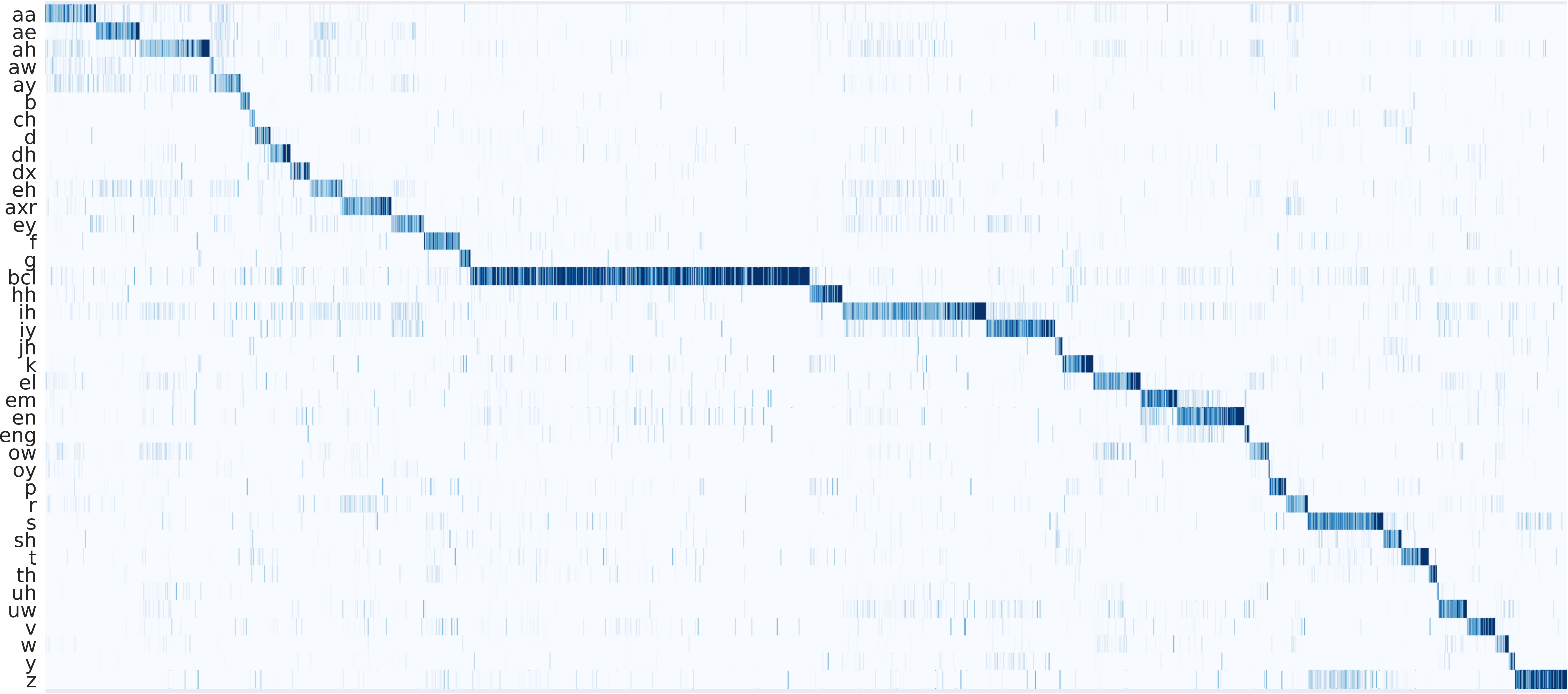}
\caption{
Visualization of the co-occurrence between discrete latent speech representations and phonemes.
We plot the conditional probability $P(phoneme|\zq_t)$ on TIMIT train data.
The y-axis shows the collapsed 39 classes of phonemes and the x-axis is over the different discrete latents.
}
\label{fig:token2phone_distribution}
\end{figure}

\autoref{fig:token2phone_distribution} plots $P(phoneme|\zq_t)$ and shows that many discrete latents appear to specialize in specific phonetic sounds.
The silence phoneme (bcl) represents 22\% of all human annotated speech data and is therefore also modeled by many different latents.

\newpage
\section{Speech Recognition Error Analysis}
\label{app:analysis_errors}

In this section we study the most common errors our models make when fine-tuned on different amounts of labeled data (\autoref{tbl:errors}). 
We also show transcriptions of a few relatively challenging utterances from the dev-clean subset of Librispeech (\autoref{tbl:examples}).

We consider models with no lexicon or no language model decoding, marked None in~\autoref{tab:librilight_full}:
Larger capacity decreases error rates: \wvppbig{} on \librisz{} improves the word error rate on dev-clean from 46.1 to 43 compared to \wvppbase{}.
Increasing the amount of unlabeled training data further decreases the error rate to 33.8 for \wvppbig{} on \librisz{}.

In the ten minute labeled data setup, the model is still able to recognize basic units of speech: \autoref{tbl:errors} shows that most errors are around spelling of words, e.g., omitting silent characters such as \emph{could} $\rightarrow$ \emph{coud}, \emph{know} $\rightarrow$ \emph{now}, or ignoring repeated letters such as \emph{still} $\rightarrow$ \emph{stil}, \emph{little} $\rightarrow$ \emph{litle}. 
The \wvppbig{} \voxsz{} model achieves WER 38.3 on dev-clean and adding a Transformer language model enables to choose more likely pronunciations during the search and gives a large WER improvement to 5.0.

The ten minute models without lexicon and language model tend to spell words phonetically and omit repeated letters, e.g., \emph{will} $\rightarrow$ \emph{wil} (\autoref{tbl:errors}).
Spelling errors decrease with more labeled data: 
with one hour of labeled data, slightly less common words move into the list of the most frequent errors, e.g., \emph{heaven} and \emph{food} are spelled phonetically. 
At ten hours, top errors include articles, e.g., \emph{a}, \emph{the} which are a common source of errors in speech recognition in general. 
There are also alternative spellings, \emph{color} vs. \emph{colour} as well as relatively rare words including person names, still spelled phonetically, e.g., \emph{phoebe} $\rightarrow$ \emph{feeby}. 

At 100 hours, person names dominate the most frequent errors: 
\emph{phoebe} $\rightarrow$ \emph{phebe}, along with incorrect spacing \emph{anyone} $\rightarrow$ \emph{any one}, \emph{awhile} $\rightarrow$ \emph{a while}. 
Finally at 960 hours the word error rate falls to 2\% and top errors are mostly articles, incorrect splits, and some very rare words or names such as \emph{deucalion} or \emph{gryce}.

The “from scratch” 960 hour model has a similar word error rate as the 100 hour pre-trained model and displays a similar pattern of errors.

The pre-trained speech representations can be easily adapted to recognize specific sounds while fine-tuning grounds these representations to the actual spelling.

\begin{table}
% \begin{adjustwidth}{-4em}{}
\caption{%
Top word errors for models trained on 10m, 1h and 10h, 100h, 960h of labeled data and decoded on the Librispeech dev-clean subset without a language model or lexicon (see~\autoref{tab:librilight_full} and \autoref{tab:librispeech_full} - None). In brackets is the total number of occurrences of each error.
}
\label{tbl:errors}
\centering 
% \resizebox{\textwidth}{!}{%
\begin{tabular}{l|l|l}
\toprule
10m \wvppbig{} \voxsz{} & 1h \wvppbig{} \voxsz{} & 10h \wvppbig{} \voxsz{} \\
\midrule
all $\rightarrow$ al (181) &too $\rightarrow$ to (26) &in $\rightarrow$ and (15) \\
are $\rightarrow$ ar (115) &until $\rightarrow$ untill (24) &a $\rightarrow$ the (11) \\
will $\rightarrow$ wil (100) &new $\rightarrow$ knew (22) &o $\rightarrow$ oh (10) \\
you $\rightarrow$ yo (90) &door $\rightarrow$ dor (18) &and $\rightarrow$ in (9) \\
one $\rightarrow$ on (89) &says $\rightarrow$ sais (18) &mode $\rightarrow$ mod (9) \\
two $\rightarrow$ to (81) &soul $\rightarrow$ sol (17) &ursus $\rightarrow$ ersus (9) \\
well $\rightarrow$ wel (80) &bread $\rightarrow$ bred (16) &tom $\rightarrow$ tome (8) \\
been $\rightarrow$ ben (73) &poor $\rightarrow$ pore (16) &randal $\rightarrow$ randol (7) \\
upon $\rightarrow$ apon (73) &a $\rightarrow$ the (13) &the $\rightarrow$ a (7) \\
good $\rightarrow$ god (67) &either $\rightarrow$ ither (13) &color $\rightarrow$ colour (6) \\
see $\rightarrow$ se (66) &food $\rightarrow$ fud (13) &flour $\rightarrow$ flower (6) \\
we $\rightarrow$ whe (60) &doubt $\rightarrow$ dout (12) &phoebe $\rightarrow$ feeby (6) \\
little $\rightarrow$ litle (54) &earth $\rightarrow$ erth (12) &an $\rightarrow$ and (5) \\
great $\rightarrow$ grate (53) &led $\rightarrow$ lead (12) &cucumbers $\rightarrow$ cucombers (5) \\
your $\rightarrow$ yor (53) &sea $\rightarrow$ see (12) &egg $\rightarrow$ eg (5) \\
could $\rightarrow$ coud (51) &thee $\rightarrow$ the (12) &macklewain $\rightarrow$ macklewaine (5) \\
here $\rightarrow$ hear (51) &tom $\rightarrow$ tome (12) &magpie $\rightarrow$ magpi (5) \\
know $\rightarrow$ now (45) &add $\rightarrow$ ad (11) &milner $\rightarrow$ millner (5) \\
there $\rightarrow$ ther (45) &good $\rightarrow$ god (11) &stacy $\rightarrow$ staci (5) \\
three $\rightarrow$ thre (45) &heaven $\rightarrow$ heven (11) &trevelyan $\rightarrow$ trevellion (5) \\
still $\rightarrow$ stil (42) &mary $\rightarrow$ marry (11) &verloc $\rightarrow$ verlock (5) \\
off $\rightarrow$ of (40) &randal $\rightarrow$ randel (11) &ann $\rightarrow$ an (4) \\
don't $\rightarrow$ dont (37) &answered $\rightarrow$ ansered (10) &anyone $\rightarrow$ one (4) \\
shall $\rightarrow$ shal (36) &blood $\rightarrow$ blod (10) &apartment $\rightarrow$ appartment (4) \\
little $\rightarrow$ litl (35) &bozzle $\rightarrow$ bosel (10) &basin $\rightarrow$ bason (4) \\
\bottomrule
\midrule
100h \wvppbig{} \voxsz{} & 960h \wvppbig{} \voxsz{} & 960h \wvppbig{} from scratch  \\
\midrule
 a $\rightarrow$ the (13) & a $\rightarrow$ the (12) & and $\rightarrow$ in (20) \\
 and $\rightarrow$ in (10) & and $\rightarrow$ in (9) & a $\rightarrow$ the (16) \\
 in $\rightarrow$ and (10) & macklewain $\rightarrow$ mackelwaine (7) & in $\rightarrow$ and (13) \\
 o $\rightarrow$ oh (8) & in $\rightarrow$ and (6) & the $\rightarrow$ a (10) \\
 minnetaki $\rightarrow$ minnitaki (7) & o $\rightarrow$ oh (6) & in $\rightarrow$ an (8) \\
 randal $\rightarrow$ randall (7) & bozzle $\rightarrow$ bosell (5) & and $\rightarrow$ an (5) \\
 christie $\rightarrow$ cristy (6) & criss $\rightarrow$ chris (5) & clarke $\rightarrow$ clark (4) \\
 macklewain $\rightarrow$ mackelwane (6) & bozzle $\rightarrow$ bosel (4) & grethel $\rightarrow$ gretel (4) \\
 randal $\rightarrow$ randoll (6) & clarke $\rightarrow$ clark (4) & macklewain $\rightarrow$ mackelwaine (4) \\
 bozzle $\rightarrow$ bosall (5) & colored $\rightarrow$ coloured (4) & this $\rightarrow$ the (4) \\
 kaliko $\rightarrow$ calico (5) & grethel $\rightarrow$ gretel (4) & an $\rightarrow$ and (3) \\
 trevelyan $\rightarrow$ trevelian (5) & lige $\rightarrow$ lyge (4) & anyone $\rightarrow$ one (3) \\
 an $\rightarrow$ and (4) & the $\rightarrow$ a (4) & bozzle $\rightarrow$ basell (3) \\
 and $\rightarrow$ an (4) & and $\rightarrow$ an (3) & buns $\rightarrow$ bunds (3) \\
 anyone $\rightarrow$ one (4) & ann $\rightarrow$ marianne (3) & carrie $\rightarrow$ carry (3) \\
 bozzle $\rightarrow$ bozall (4) & butte $\rightarrow$ bute (3) & criss $\rightarrow$ chris (3) \\
 clarke $\rightarrow$ clark (4) & color $\rightarrow$ colour (3) & he's $\rightarrow$ is (3) \\
 gryce $\rightarrow$ grice (4) & deucalion $\rightarrow$ ducalion (3) & his $\rightarrow$ is (3) \\
 i'm $\rightarrow$ am (4) & forcemeat $\rightarrow$ meat (3) & honor $\rightarrow$ honour (3) \\
 in $\rightarrow$ ind (4) & gryce $\rightarrow$ grice (3) & lattimer $\rightarrow$ latimer (3) \\
 letty $\rightarrow$ lettie (4) & honor $\rightarrow$ honour (3) & millet $\rightarrow$ mellet (3) \\
 phoebe $\rightarrow$ phebe (4) & kearny $\rightarrow$ kirney (3) & pyncheon $\rightarrow$ pension (3) \\
 the $\rightarrow$ a (4) & nuova $\rightarrow$ noiva (3) & tad $\rightarrow$ ted (3) \\
 ann $\rightarrow$ anne (3) & thing $\rightarrow$ anything (3) & thing $\rightarrow$ anything (3) \\
 awhile $\rightarrow$ while (3) & this $\rightarrow$ the (3) & trevelyan $\rightarrow$ trevelian (3) \\
\bottomrule
\end{tabular}
% }
% \end{adjustwidth}
\end{table}

\begin{table}
% \begin{adjustwidth}{-5em}{}
\caption{%
Examples of transcription of selected utterances from the dev-clean subset by various models without a language model or lexicon. Capitalized words indicate errors.
}
\label{tbl:examples}
\centering 
\resizebox{\textwidth}{!}{%
\begin{tabular}{l|l}
\toprule
Model & Transcription \\
\midrule
Reference & i'm mister christopher from london \\
10m \voxsz{} & IM mister CRESTIFER FROME LUNDEN \\
1h \voxsz{} & IM mister CRISTIFFHER from LOUNDEN \\
10h \voxsz{} & i'm mister CHRYSTEPHER from london \\
100h \voxsz{} & i'm mister christopher from london \\
960h \voxsz{} & i'm mister christopher from london \\
960h scratch & I MISSTER christopher from london \\
\midrule
Reference & il popolo e una bestia \\
10m \voxsz{} & ILPOPULAR ONABESTIA \\
1h \voxsz{} & O POPOLAONABASTIA \\
10h \voxsz{} & U POPULAONABASTIAR \\
100h \voxsz{} & O POPALOON A BASTYA \\
960h \voxsz{} & YOU'LL POP A LAWYE ON A BAISTYE \\
960h scratch & OL POPALOY ON ABESTIA \\
\midrule
Reference & he smelt the nutty aroma of the spirit \\
10m \voxsz{} & he SMELTD the NUDY  aroma of the spirit \\
1h \voxsz{} & he SMELTD the NUDDY ARROMA of the spirit \\
10h \voxsz{} & he smelt the NUDDY ERROMA of the spirit \\
100h \voxsz{} & he smelt the NUDDY aroma of the spirit \\
960h \voxsz{} & he smelt the NUTTIE aroma of the spirit \\
960h scratch & he smelt the nutty EROMA of the spirit \\
\midrule
Reference & phoebe merely glanced at it and gave it back \\
10m \voxsz{} & FEABY  MEARLY glanced at it and gave it BAK \\
1h \voxsz{} & FIEABY merely glanced at it and gave it back \\
10h \voxsz{} & FEEBY  merely glanced at it and gave it back \\
100h \voxsz{} & BEBE   merely glanced at it and gave it back \\
960h \voxsz{} & phoebe merely glanced at it and gave it back \\
960h scratch & phoebe merely glanced at it and gave it back \\
\midrule
Reference & sauterne is a white bordeaux a strong luscious wine the best known varieties being \\
10m \voxsz{} & SULTERIN is a white BORDOE   a strong LUCHOUS  WIN  the best NOWN  VERIATYS  being \\
1h \voxsz{} & CLTEREN  is a white BORDO    a strong LUCHIOUS wine the best known VERIETIES being \\
10h \voxsz{} & SOTERN   is a white BOURDO   a strong LUCIOUS  wine the best known VORIETIES being \\
100h \voxsz{} & SOTERN   is a white BORDAUX  a strong LUCIOUS  wine the best known varieties being \\
960h \voxsz{} & SOTERN   is a white bordeaux a strong luscious wine the best known varieties being \\
960h scratch & SOTERAN  is a white bordeaux a strong luscious wine the best known varieties being \\
\midrule
Reference & i happen to have mac connell's box for tonight or there'd be no chance of our getting places \\
10m \voxsz{} & i HAPEND to have MECONALES BOXS for TONIT ORE THIRLD  be no chance of OR  GETING  places \\
1h \voxsz{} & i happen to have MACCONNEL'S BOCXS for tonight or TE'ELD  be no chance of our getting places \\
10h \voxsz{} & i HAPPENED to have MUKONNEL'S box for tonight or THERED  be no chance of our getting places \\
100h \voxsz{} & i HAPPENED to have MC  CONNEL'S  box for TO NIGHT or there'd be no chance of our getting places \\
960h \voxsz{} & i happen to have MC  CONALL'S  box for TO NIGHT   or there'd be no chance of our getting places \\
960h scratch & i HAPPENE to have MACONEL'S box for TO NIGHT   or there'd be no chance of our getting places \\
\bottomrule
\end{tabular}}
% \end{adjustwidth}
\end{table}

\newpage
\section{Ablations}
\label{app:ablations}

\autoref{tbl:ablations} ablates various hyperparameter choices of our architecture. The setup for the baseline model is described in~\autoref{sec:ablations}. 
First, we tried to improve the continuous input and continuous target model (\autoref{sec:ablations}) by adding an MLP on top of the continuous target representation and we also tried to use a separate set of encoder parameters for the representations used as input and targets (Separate encoders). 
Both did not lead to meaningful improvements.

Increasing the receptive field size from 25ms to 30ms had little effect.
Setting the diversity penalty weight ($\alpha{}$) too low results in lower codebook usage and lower performance.
Setting it too high leads to slight instability.
% Removing the L2 penalty over the feature encoder outputs decreases performance ($\beta=0$).
Doubling the number of relative positional embeddings to 256 also did not help.
Stopping gradients from the quantizer to the encoder shows that the encoder requires training signal from the quantizer as well.

\begin{table}[h!]
\caption{%
Ablation of various hyper-parmeter choices. We report average WER and standard deviation on combined dev-clean/other of \libri{} for three seeds of training.
}
\label{tbl:ablations}
\centering
\begin{tabular}{lrr}
\toprule
{} & avg. WER &  std. \\
\midrule
Baseline ($p=0.075$, $\alpha=0.1$) & 7.97 & 0.02 \\
\midrule
Continuous inputs, continuous targets & 8.58 & 0.08 \\
+ MLP on targets & 8.51 & 0.05 \\
+ Separate encoders & 8.90 & 0.01 \\
\midrule
receptive field 30ms & 7.99 & 0.06 \\
\midrule
diversity penalty \\
$\alpha=0$ & 8.48 & 0.08 \\
$\alpha=0.05$ & 8.34 & 0.08 \\
$\alpha=0.2$ & 8.58 & 0.45 \\
\midrule
% Feature L2 penalty 
% $\beta=0$ & 8.38 & 0.07 \\
\midrule
Conv pos emb, kernel 256 & 8.14 & 0.05 \\
\midrule
No gradient to encoder from quantizer & 8.41 & 0.08 \\
\midrule
Negatives \\
$K=200$ same utterance & 8.12 & 0.05 \\
$K=50$ same utterance + $K=50$ from batch & 8.79 & 0.06 \\
\midrule
Sample negatives from any time step & 8.07 & 0.02 \\
\midrule
No Gumbel noise & 8.73 & 0.42 \\
\midrule
Codebook \\
G=4, V=18 & 9.02 & 0.38 \\
G=8, V=8 & 8.13 & 0.07 \\
\midrule
Predict exactly $U$ time steps from edges \\
$U=1$ & 9.53 & 0.91 \\
$U=5$ & 8.19 & 0.07 \\
$U=10$ & 8.07 & 0.07 \\
$U=15$ & 7.89 & 0.10 \\
$U=20$ & 7.90 & 0.01 \\
\bottomrule
\end{tabular}
\end{table}

Next, increasing the number of negatives did not result in better performance ($K=200$) and sampling negatives from the entire batch of utterances hurt performance, likely because candidates from other utterances are easy to distinguish.
Sampling negatives from any time step in the utterance, masked or unmasked, does not help and is more computationally expensive.
Gumbel noise is important and increasing the number of codebooks did not result in better performance.

We also investigated predicting only time steps immediately next to the last unmasked time step for each span. 
This enables to better control the difficulty of the pre-training task.
Given the leftmost or rightmost unmasked time step next to a masked span, we compute the contrastive loss only for the first $U$ masked time steps next to these unsmasked spans.
Predicting only up to one time step performs poorly because there is little training signal from each utterance and predicting more time steps performs better but does not significantly outperform predicting all masked time steps.
Increasing the number of training updates helps but this increases training time.

\end{document}

%% file: macro.tex
\newcommand{\wvpp}{wav2vec 2.0}
\newcommand{\wvppbase}{\textsc{Base}}
\newcommand{\wvppbig}{\textsc{Large}}
\newcommand{\vox}{LibriVox}
\newcommand{\libri}{Librispeech}
\newcommand{\libril}{Libri-light}
\newcommand{\voxsz}{LV-60k}
\newcommand{\librisz}{LS-960}
\newcommand{\libriunsz}{LS-860}

\newcommand{\Inp}{\mathcal{X}}
\newcommand{\Feat}{\mathcal{Z}}
\newcommand{\QFeat}{\mathcal{Q}}
\newcommand{\Context}{\mathcal{C}}
\newcommand{\R}{\mathbb{R}}

\newcommand{\ze}{\mathbf{z}}
\newcommand{\zq}{\mathbf{q}}
\newcommand{\zqt}{\mathbf{\tilde{q}}}
\newcommand{\cc}{\mathbf{c}}

\newcommand{\lll}{\mathbf{l}}
\newcommand{\Lmlm}{\mathcal{L}_{m}}
\newcommand{\Ld}{\mathcal{L}_{d}}

\newcommand{\Enot}[1]{\num[exponent-product = \times]{#1}}